\documentclass[11pt]{article}

\usepackage[final]{acl}

\usepackage{times}
\usepackage{latexsym}
\usepackage[dvipsnames]{xcolor}
\usepackage{multicol}
\usepackage{multirow}
\usepackage{booktabs}
\usepackage{amssymb}
\usepackage{booktabs}   % \toprule, \midrule, \bottomrule
\usepackage{array}      % custom p{} column formatting
\usepackage[table]{xcolor} % \rowcolor and gray!10
\usepackage[T1]{fontenc}
\usepackage{lipsum}
\usepackage{arydshln}

\usepackage[utf8]{inputenc}
\usepackage[T1]{fontenc}
\usepackage{microtype}

\usepackage{inconsolata}

\usepackage{graphicx}
\usepackage{enumitem}
\usepackage[normalem]{ulem} % for strikethrough
\usepackage{xcolor} % for colored text
\usepackage{booktabs}
\usepackage{url}
\usepackage[most]{tcolorbox}
\usepackage{fancyvrb}
\usepackage{pifont} % for checkmark
\usepackage{graphicx} % for \resizebox
\usepackage{soul} % for the command \hl

\usepackage{cuted}   % provides strip environment

\title{Halluscoring 2026: The First Shared Task on LLM Hallucination Detection and Answer Verification.}

\author{%
\normalfont
Aisha Alansari$^{1}$,
Abdessalam Bouchekif$^{2}$,
Ahmed Hasanaath$^{1}$,
Salah Eddine Bekhouche$^{3}$,\\
Malak Alkhorasani$^{4}$,
Mohammed-En-Nadhir Zighem$^{2}$,
Saad Ezzini$^{1}$,
Hichem Telli$^{5}$,\\
Hend Al-Khalifa$^{8}$,
Muhammad Abdul-Mageed$^{6}$,
Hadid Abdenour$^{7}$,
Hamzah Luqman$^{1}$\\ [6pt]
{\normalfont\footnotesize
\begin{tabular}{@{}l@{\hspace{1.5em}}l@{}}
$^{1}$ King Fahd University of Petroleum and Minerals, Saudi Arabia &
$^{5}$ University of Biskra, Algeria \\
$^{2}$ Hamad Bin Khalifa University, Qatar &
$^{6}$ University of British Columbia, Canada \\
$^{3}$ University of the Basque Country, Spain &
$^{7}$ Universiti Malaysia Kelantan, Malaysia \\
$^{4}$ Imam Abdulrahman bin Faisal University, Saudi Arabia &
$^{8}$ King Saud University, Saudi Arabia \\
[3pt]
\multicolumn{2}{c}{
\texttt{\{aisha.ansari, g202302610, hluqman\}@kfupm.edu.sa}
}
\end{tabular}
}
}

\begin{document}
\maketitle

% \begin{figure*}[t!]
%     \centering
%     \includegraphics[width=\linewidth]{Figures/Overview.pdf}
%     \caption{Description of the HalluScoring 2026 shared task.}
%     \label{fig:overview}
% \end{figure*}

\setlength{\stripsep}{8pt}

\begin{strip}
\centering\includegraphics[width=\linewidth]{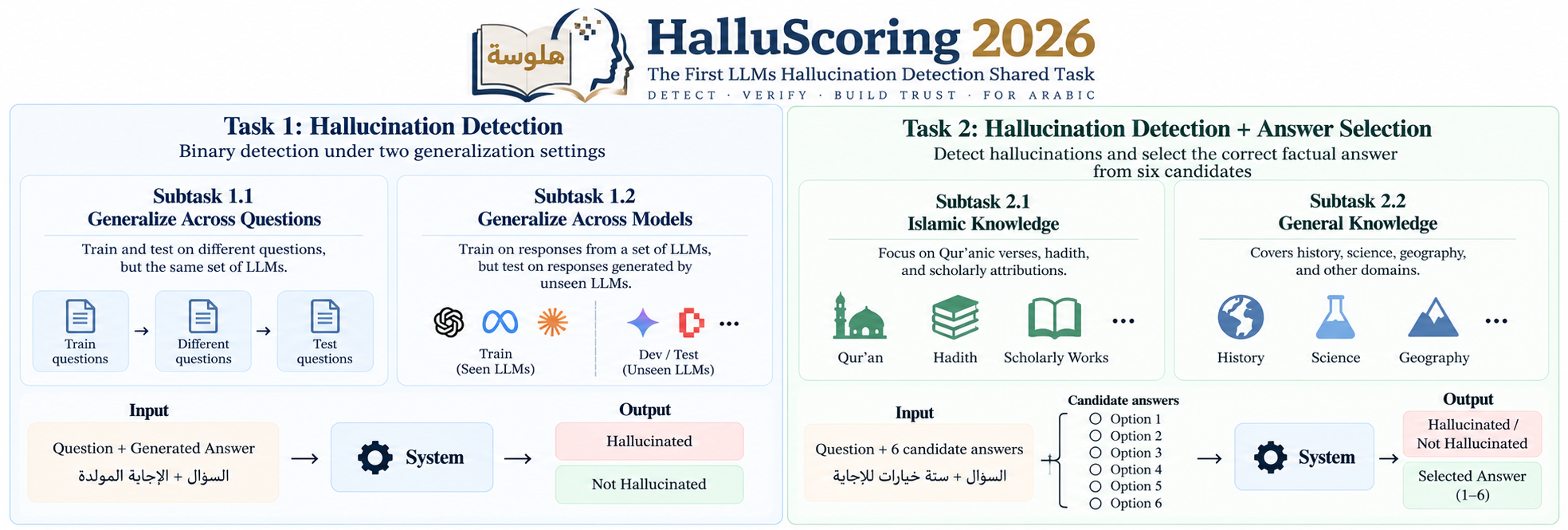}
\captionof{figure}{Overview of the \textit{HalluScoring 2026} shared task. Task~1 evaluates binary hallucination detection under two generalization settings: unseen questions (Subtask~1.1) and unseen LLMs (Subtask~1.2). Task~2 extends the evaluation to two stages, first detecting if a generated response is hallucinated and then selecting the correct factual answer from six candidate options, covering Islamic knowledge (Subtask~2.1) and general knowledge (Subtask~2.2).}
    \label{fig:overview}
\end{strip}

\begin{abstract}

We present \textit{HalluScoring 2026}, a shared task for evaluating hallucination detection and factual verification in Arabic question answering under challenging generalization settings. Its four subtasks are organized into two tasks. Task~1 evaluates binary hallucination detection, considering generalization to unseen questions (Subtask~1.1) and responses generated by unseen LLMs (Subtask~1.2). Task~2 extends the evaluation beyond detection by requiring the systems to additionally identify the correct factual answer from six related candidates, covering Islamic knowledge (Subtask~2.1) and general knowledge (Subtask~2.2). The shared task is based on two Arabic datasets: HalluScore and HalluTruthQA. A total of 13 teams participated in the shared task, ten of which submitted system description papers. The results of Task~1 demonstrate that hallucination detection remains challenging. On the Task~1 test sets, the top-ranked systems achieved AUC-ROC scores of 0.7717 for Subtask~1.1 (REGLAT) and 0.7670 for Subtask~1.2 (NAMAA). Under assisted evaluation, the highest combined detection and answer-selection scores for Subtasks~2.1 and~2.2 were 0.8824 and 0.8565, respectively.

\end{abstract}

\section{Introduction}

Large language models (LLMs) are increasingly being used to generate and process Arabic content in a wide range of applications, including question answering (QA), information retrieval, and summarization \cite{hasanaath2025arareasoner,alzubaidi2025evaluating,mashabi2026survey}. Recent Arabic-specific and multilingual LLMs can produce fluent and contextually appropriate Arabic text, supporting their use in both general-purpose and domain-specific applications \cite{alam2024llms,abdaljalil2025evaluating,bouchekif-etal-2025-assessing}. However, fluent responses are not necessarily factually reliable. LLMs may present fabricated, unsupported, or factually incorrect claims as true; we refer to such claims as hallucinations \citep{maynez2020faithfulness,ji2023survey}. Such hallucinations may involve entities, dates, numerical values, quotations, source attributions, or supporting explanations \cite{li2023inference,alansari2025arahallueval,bouchekif2026hallutruthqa}.

% Large language models (LLMs) can generate fluent and convincing responses, but fluency does not guarantee factual correctness. Despite their strong capabilities, LLMs can still produce hallucinated content that is factually incorrect, fabricated, unsupported, or inconsistent with available evidence \citep{maynez2020faithfulness,ji2023survey, bouchekif-etal-2025-assessing}. Such errors may involve entities, dates, numerical values, quotations, source attributions, or supporting explanations \citep{alansari2025arahallueval,bouchekif2026hallutruthqa}. Importantly, a response may contain the correct main answer while still including incorrect evidence, attribution, or supporting details.

Hallucinations may arise from incomplete or noisy training data, limited coverage of domain-specific knowledge, ambiguous prompts, or insufficient contextual grounding. They may also be influenced by LLMs' next-token prediction objective, which optimizes for likely continuations without explicitly verifying the factual correctness of each generated claim \citep{huang2025survey,alansari2026large}. This challenge is particularly relevant to Arabic, where high-quality, diverse, and domain-specific evaluation resources remain comparatively limited.

The evaluation of hallucinations in Arabic remains underexplored, particularly with respect to generalization beyond the conditions observed during training. A detector that performs well on questions similar to those observed during training may not generalize to new questions \cite{alansari2026crosshallu}. Likewise, a system trained on responses generated by a fixed set of LLMs may capture generator-specific characteristics rather than general hallucination signals, potentially limiting its ability to generalize to responses produced by unseen models \cite{wu2025survey}. Moreover, correctly detecting a hallucination does not necessarily imply that a system can identify the correct factual information \cite{bouchekif2026hallutruthqa}. These challenges motivate the evaluation of both the robustness of hallucination detection and the ability to identify the correct answer.

To address these challenges, we present \textit{HalluScoring 2026}, a shared task on hallucination detection and factual verification in Arabic QA. As shown in Figure \ref{fig:overview}, the shared task aims to evaluate the robustness of hallucination detection under two complementary generalization settings: Task~1 evaluates binary hallucination detection on unseen questions (Subtask~1.1) and on responses from unseen LLMs to unseen questions (Subtask~1.2). Task~2 requires systems to detect hallucinations and select the correct answer from six candidates.  Subtask~2.1 focuses on Islamic knowledge, while Subtask~2.2 covers broader domains, including history, science, and geography. The shared task is based on two carefully annotated Arabic resources, \textit{HalluScore} \cite{alansari2026halluscore} and \textit{HalluTruthQA} \citep{bouchekif2026hallutruthqa,bekhouche2026hallutruthqa-4k}, enabling evaluation across multiple domains and model generators using expert-verified factual information. We also provide a comparative evaluation of participating systems, highlighting methodological approaches, performance trends, and remaining challenges across different evaluation settings.

\section{Literature Review}\label{sec:lit}

Hallucination in LLMs has motivated several shared tasks aimed at developing standardized evaluation settings for hallucination detection. SemEval-2024 Task 6, \textit{SHROOM} \cite{mickus2024semeval}, focused on detecting fluent but semantically incorrect outputs in machine translation, paraphrase generation, and definition modeling, considering both model-aware and model-agnostic settings. Its multilingual successor, SemEval-2025 Task~3, \textit{Mu-SHROOM} \cite{vazquez2025semeval}, extended this work to 14 languages, including Arabic, and evaluated the localization of hallucinated spans in LLM outputs. These shared tasks established standardized settings for evaluating hallucination detection in language and generation tasks.

Arabic research on hallucination has also progressed through both shared tasks and dedicated datasets. The \textit{Halwasa} shared task \cite{mubarak2024halwasa} addressed hallucination detection in Arabic LLM-generated content, building on a dataset of 10,000 Arabic sentences annotated for factual and linguistic correctness. \textit{IslamicEval 2025} \cite{mubarak2025islamiceval} subsequently targeted hallucinations within the specialized domain of Islamic content, including the detection and correction of hallucinated Qur'anic verses and Hadiths grounded in authoritative sources. Beyond shared tasks, \textit{AraHalluEval} \cite{alansari2025arahallueval} introduced fine-grained hallucination annotations for Arabic generative QA and summarization, distinguishing different factuality and faithfulness errors. 

\paragraph{Positioning of HalluScoring 2026.}
\textit{HalluScoring 2026} extends Arabic hallucination shared tasks by
evaluating both the robustness of hallucination detection and the ability to
recover correct factual information. Building on \textit{HalluScore} \cite{alansari2026halluscore} and \textit{HalluTruthQA} \cite{bouchekif2026hallutruthqa,bekhouche2026hallutruthqa-4k}, the shared task evaluates generalization to unseen questions and LLMs, as well as joint hallucination detection and correct-answer selection. Moreover, the evaluation covers both Islamic and general-knowledge domains, with the latter spanning areas such as history, science, and geography. Reliable hallucination detection is particularly important in these settings because factually incorrect responses can misrepresent religious teachings, historical events, scientific facts, or geographical information, potentially misleading users who rely on LLMs as sources of information.

\section{HalluScoring 2026}
\label{sec:HalluScoring}

\subsection{Task 1: Hallucination Detection}
The goal of Task 1 is to detect hallucinations in Arabic QA in diverse domains. Given a question and one or more model-generated responses, the objective is to automatically determine whether each response contains hallucinated content. The questions span multiple domains and topics, exposing models to different types of knowledge and varying levels of factual specificity, which can make generated responses particularly prone to hallucination. To encourage the development of robust detection methods that generalize across different questions and LLMs, the shared task is organized into two complementary subtasks.

\noindent
\textbf{Subtask 1.1: Generalize Across Questions}
In this subtask, participants are required to develop hallucination detection systems that can generalize across different questions. The training and test splits are constructed such that all models appear in both sets, but the questions are disjoint. Each instance consists of a question and a generated response, in which the system must predict whether the response is hallucinated or non-hallucinated. 

\noindent
\textbf{Subtask 1.2: Generalize Across Models}
This subtask evaluates the ability of the proposed hallucination detection techniques to generalize across unseen models. The dataset is split such that the training set contains responses generated by one subset of LLMs, while the test set contains responses from different LLMs that are entirely unseen during training. The questions are also partitioned between the training and test sets to evaluate generalization across both models and questions. The goal remains to classify hallucinated versus non-hallucinated outputs. This setting is particularly challenging, as it requires learning model-invariant hallucination signals that transfer across different architectures.

\subsection{Task 2: From Hallucination Detection to Truth}

The goal of Task 2 is to evaluate whether a system can both detect hallucinations and identify the correct information. Given an Arabic question and a generated answer, the system must first determine whether the answer contains a hallucination. Then, the correct answer must be selected from six candidate options, including a correct answer and five plausible distractors. The candidate options are designed to be close to one another, making it more challenging to distinguish correct from incorrect answers. Task 2 consists of two subtasks.

\noindent
\textbf{Subtask 2.1: Islamic Knowledge.}
This subtask focuses on Islamic knowledge. The questions cover topics such as Prophetic biography, Islamic law, religious concepts, and source-sensitive information, including Qur'anic verses, Hadith, scholarly opinions, and source attributions. This domain is particularly challenging for hallucination detection because a generated answer may provide the correct main answer while still containing incorrect details, such as a verse number, a quotation, an attribution, or a supporting statement.

\noindent
\textbf{Subtask 2.2: General Culture.}
This subtask focuses on general knowledge across a wide range of topics, including history, geography, science, culture, and Islamic knowledge. It evaluates the system's ability to detect factual errors and identify the correct answer across diverse domains. The questions may contain plausible but incorrect information, including errors in dates, entities, locations, events, names, numerical facts, or other factual details, requiring the system to carefully distinguish between accurate and misleading information.
Subtask~2.1 focuses on Islamic knowledge, while Subtask~2.2 extends the evaluation to three additional domains: history, science, and geography.

\subsection{Task Setup}

\textit{HalluScoring 2026} was organized by hosting four separate CodaBench competitions, one for each subtask. This separation allowed each task to have its own data format, submission pipeline, and evaluation procedure. The training and development sets for all tasks were released in 16th May 2026. For Task 1, the training and development data were provided in \texttt{Excel} format, whereas for Task~2, they were provided in \texttt{JSON} format. Appendix \ref{sec:data_samples} illustrates some samples given to the participants.

Participants used these released splits to develop and validate their systems before submission. To preserve the integrity of the final evaluation, we did not publicly release the test sets to participants. Instead, we provided a Jupyter Notebook template that specified the required procedures for model loading, inference, and output. Participants were asked to implement their systems using this template and submit the completed notebook to the organizers. The organizers executed the submitted notebooks on the hidden test sets.

%write the total and how many submitted description papers.

\subsection{Shared Task Restrictions}
To ensure a fair and controlled evaluation setting, we provided clear restrictions on participating systems via the CodaBench platform. All submissions were required to be based on a fine-tuned generative language model whose base model is open-source or openly available for research use. We restricted submitted models to a maximum of 13 billion parameters to maintain comparable computational settings between participants. During evaluation, participants were not allowed to use retrieval-augmented generation (RAG), external search tools, or other retrieval-based systems. This restriction was introduced to ensure that performance primarily reflects the capabilities of the submitted model rather than its access to external information during inference. However, participants were allowed to use external datasets for model fine-tuning and were requested to disclose any external datasets. These constraints were designed to provide a standardized evaluation setting while still allowing teams flexibility in model selection, training data, and fine-tuning strategies.
\section{Datasets and Evaluation Metrics}

\subsection{HalluScore Dataset}
The \textit{HalluScore} \cite{alansari2026halluscore} dataset is proposed for the evaluation, detection, and mitigation of Arabic hallucination in LLMs. It is a structured Arabic QA benchmark comprising 827 carefully curated QA pairs across multiple domains, including health, science, finance, religion, and geography. The dataset is constructed through a rigorous pipeline involving quality assurance, filtering for clarity and factual validity, and model-driven selection to retain questions that consistently trigger hallucinations. For each question, responses were generated by 17 Arabic, multilingual, and reasoning LLMs and manually annotated as either hallucinated or non-hallucinated, resulting in a total of $14{,}059$ annotated QA-response instances. For the shared task, only the information required for the corresponding prediction tasks was provided to participants, while the additional annotations and metadata available in the original \textit{HalluScore} benchmark were not included as prediction targets. The test sets and their corresponding labels were kept hidden from participants and were used exclusively by the organizers for the final evaluation. The annotation guidelines are provided in Appendix~\ref{sec:HalluScoringAnnotation}.

\subsection{HalluTruthQA Dataset}
\label{HalluTruthQA_Dataset}
HalluTruthQA~\cite{bekhouche2026hallutruthqa-4k} contains 4,000 Arabic
question--answering instances covering four  domains:
Islamic knowledge, history, science, and geography. The corpus is balanced,
with 1,000 instances per domain. Each instance includes a question,
a model-generated answer, an expert-verified reference answer, a binary
hallucination label, and six candidate answers consisting of one correct
answer and five plausible distractors. The candidate answers were manually
constructed to be semantically related and stylistically similar, reducing
the possibility of identifying the correct answer through superficial cues. 
The model-generated answers were produced using
\texttt{QCRI/Fanar-1-9B-Instruct}~\citep{fanarllm2025}
with a maximum of 1,024 new tokens, a temperature of 0.0,
top-$p$ set to 1.0, and a batch size of 4, using
\texttt{bfloat16} precision.\footnote{Experiments were run on an
NVIDIA RTX 5880 Ada GPU with 48\,GB of memory.}
\\
Beyond the HalluScoring task's objectives of binary hallucination detection and correct-option selection, the original HalluTruthQA corpus provides richer fine-grained annotations.  In particular, hallucinated responses include character-level erroneous spans and human-written explanations describing why each span is incorrect. This supports error localization, explanation, and fine-grained analysis beyond the official shared-task objectives. The annotation guidelines are provided in Appendix~\ref{sec:HalluTruthQAAnnotation}.

\subsection{Data Splits}
The numbers of LLMs, questions, and instances in the train, development, and test sets are reported in Table~\ref{tab:task_stats}. Subtask~1.1 and 1.2 combine \textit{HalluScore} and \textit{HalluTruthQA} to increase the number and diversity of instances available for training and evaluation. Subtask~2.1 uses the Islamic-knowledge subset of \textit{HalluTruthQA},
whereas Subtask~2.2 uses the full mixed-domain dataset, covering Islamic
knowledge, history, science, and geography.

\begin{table*}[t]
\centering

\small

\begin{tabular}{llccc|ccc|ccc}
\toprule
& &
\multicolumn{3}{c|}{\textbf{\# LLMs}} &
\multicolumn{3}{c|}{\textbf{\# Questions}} &
\multicolumn{3}{c}{\textbf{\# Instances}} \\
\cmidrule(lr){3-5}
\cmidrule(lr){6-8}
\cmidrule(lr){9-11}

\textbf{Subtask} &
\textbf{Dataset} &
\textbf{Train} &
\textbf{Dev.} &
\textbf{Test} &
\textbf{Train} &
\textbf{Dev.} &
\textbf{Test} &
\textbf{Train} &
\textbf{Dev.} &
\textbf{Test} \\
\midrule

\shortstack[l]{Subtask 1.1\\Across Questions}
& \shortstack[l]{HalluScore\\HalluTruthQA}
& 5 & 5 & 5
& 2,221 & 900 & 206
& 4,705 & 1,300 & 1,030 \\
\midrule

\shortstack[l]{Subtask 1.2\\Across Models}
& \shortstack[l]{HalluScore\\HalluTruthQA}
& 5 & 2 & 2
& 2,221 & 100 & 206
& 4,705 & 200 & 412 \\

\midrule

\shortstack[l]{Subtask 2.1\\Islamic Knowledge}
& HalluTruthQA
& 1 & 1 & 1
& 400 & 400 & 200
& 400 & 400 & 200 \\

\midrule

\shortstack[l]{Subtask 2.2\\General Culture}
& HalluTruthQA
& 1 & 1 & 1
& 1,600 & 800 & 1,600
& 1,600 & 800 &  1,600\\

\bottomrule
\end{tabular}
% \caption{Dataset statistics for the four subtasks of HalluScoring 2026.}
\caption{Dataset statistics for the four subtasks of HalluScoring 2026. For the HalluScore dataset, train set responses are generated by five models: GPT-o3, ALLaM, DeepSeek-V3, Fanar, and Claude Sonnet, whereas for HalluTruthQA, the responses are generated by Fanar.}
\label{tab:task_stats}

\end{table*}

\subsection{Evaluation Metrics}

For Task 1, we evaluate the submitted hallucination detection systems using two widely adopted metrics: area under the receiver operating characteristic curve (AUC-ROC) and macro F1-score, allowing a robust assessment of performance, particularly in the presence of class imbalance.
\\
For Task~2, evaluation is performed in two steps. First, hallucination
detection is evaluated using the macro F1-score, $F1_{\mathrm{macro}}$.
Second, the system selects the correct option among six candidates (A--F).
An instance is counted as correct for answer selection only if both the
hallucination label and the selected option are correct; otherwise, it
receives a score of zero.
\\
 Let $N_{\mathrm{both}}$ denote the number of instances with correct hallucination detection and option selection, and $N$ the total number of test instances. The answer-selection accuracy is
$
\mathrm{AnswerAcc} = \frac{N_{\mathrm{both}}}{N}.
$
The final score gives equal weight to hallucination detection and answer
selection:
\vspace{-0.3em}
\begin{equation*}
\mathrm{Score}_{\mathrm{Task2}}
=
0.5\,F1_{\mathrm{macro}}
+
0.5\,\mathrm{AnswerAcc}.
\end{equation*}
We consider two evaluation settings, \textit{blind} and \textit{assisted},
which differ in the information available during the hallucination
detection stage. In the \textit{blind} setting, the system receives only
the question and the generated answer when predicting whether the response
is hallucinated. In the \textit{assisted} setting, the six candidate
answers, which consist of one correct option and five plausible distractors,
are additionally provided during this prediction. In both settings, the
candidate answers are available for the subsequent answer-selection stage.

% The AUC-ROC metric measures the system’s ability to distinguish between hallucinated and non-hallucinated responses across all possible classification thresholds, providing a threshold-independent assessment of separability. This serves as the primary evaluation metric. Besides, the macro F1-score computes the F1-score independently for each class (hallucinated and non-hallucinated) and then averages them, ensuring equal importance is given to both classes regardless of their distribution.

\section{Shared Task Teams \& Results}\label{sec:teams_results}

%In total, the \textit{HalluScoring 2026} shared task received \textbf{10} final system submissions across its three subtasks. We received \textbf{...} submissions for Task~1.1 and \textbf{...} for Task~1.2, totaling \textbf{...} submissions across the HalluScore tasks. For Task~2, we received \textbf{...} submissions. 
In total, \textit{HalluScoring 2026} received submissions from \textbf{13 teams}, of which \textbf{10 teams} submitted system description papers. Of the participating teams, \textbf{5} participated exclusively in Task~1, \textbf{6} exclusively in Task~2, and \textbf{2} participated in both tasks.

\begin{table*}[t]
\centering
\small
\setlength{\tabcolsep}{1pt}
\renewcommand{\arraystretch}{1.08}

\begin{tabular}{
    >{\raggedright\arraybackslash}p{3.2cm}
    >{\centering\arraybackslash}p{1.4cm}
    >{\raggedright\arraybackslash}p{11.0cm}
}
\toprule
\textbf{Team} & \textbf{Subtask} & \textbf{Description} \\
\midrule

Anhnamxtanh \cite{nguyen-takahashi-2026-anhnamxtanh}
& All
& QLoRA-adapted ALLaM-7B with counterfactual cross-check recognition (C3R), using complementary TRUTH and MATCH roles, counterfactual training, option permutations, and constrained likelihood scoring. \\

\rowcolor{gray!10}
AyahVerse~\cite{rashid-2026-ayahverse}
& 1.1, 1.2, 2.2
& CAMeLBERT-based hallucination detection for Tasks~1.1--1.2 and a detection--verification pipeline combining CAMeLBERT with QLoRA-tuned AceGPT-v2-8B-Chat for Tasks~2.1--2.2. \\

DzairVerse \cite{chenini-belhadef-2026-dzairverse}
& 2.1
& QLoRA-tuned Fanar-1-9B-Instruct with a leak-free two-stage pipeline: blind hallucination detection followed by candidate-based answer selection using separate prompts. \\

\rowcolor{gray!10}
Hashtag AI \cite{chaudhuri-etal-2026-hashtag}
& 2.1, 2.2
& QLoRA-adapted Gemma 4 12B with offline knowledge-graph teacher distillation and contrastive training examples, followed by hallucination detection and candidate answer selection. \\

HelsinkiTeam 
\cite{bounab-etal-2026-helsinkiteam}
& 2.2
& LoRA-adapted ALLaM-7B-Instruct trained with approximately $30K$ augmented domain-specific and synthetic examples for joint hallucination detection and answer selection. \\

\rowcolor{gray!10}
ShadowDz \cite{kraimia-eutamene-2026-shadowdz}
& 2.1
& LoRA-tuned Fanar-1-9B-Instruct using blind hallucination detection followed by a separate candidate-selection stage conditioned on the predicted hallucination label. \\

NAMAA \cite{albalhy-etal-2026-namaa}
& 1.1, 1.2
& CAMeLBERT cross-encoder for reference-free hallucination detection, with additional experiments on knowledge distillation, contrastive learning, ensembling, calibration, and uncertainty estimation. \\

\rowcolor{gray!10}
REGLAT \cite{ashraf-etal-2026-reglat}
& 1.1, 1.2
& Soft-voting ensemble of MARBERTv2, CAMeLBERT-Mix, and ARBERT classifiers trained with Focal Loss on question--answer pairs. \\

Scalar\_NITK \cite{r-madasamy-2026-scalar}
& 1.1, 1.2
& Teacher--student knowledge-distillation framework combining an encoder ensemble and external linguistic features into a QLoRA-tuned Qwen2.5-3B-Instruct student. \\

\rowcolor{gray!10}
Vescera \cite{sellam-etal-2026-vescera}
& 2.2
& LoRA-adapted ALLaM-7B with privileged-information distillation, hidden-representation alignment, prompt paraphrasing, and option permutation for blind inference. \\

\bottomrule
\end{tabular}
\caption{Overview of the approaches adopted by teams that submitted system description papers to HalluScoring 2026.}
\label{tab:team_approaches}
\end{table*}

\subsection{Selected Participating Teams}
Table \ref{tab:team_approaches} provides an overview of the systems submitted for each subtask and summarizes the approaches adopted by the participating teams. In this section, we discuss some selected participating teams in more detail to highlight their methodological approaches in the different subtasks.

\noindent
\textbf{Anhnamxtanh} proposed counterfactual cross-check recognition (C3R), based on a QLoRA-adapted ALLaM-7B model. C3R uses two complementary roles: a TRUTH role that identifies the factually correct option and a MATCH role that selects the option best matching the response. A response is classified as non-hallucinated when the two roles agree and hallucinated otherwise. The approach also employs counterfactual training examples, option-order permutations, and constrained likelihood scoring to improve robustness.

\noindent
\textbf{AyahVerse} fine-tuned CamelBERT for binary hallucination detection using the final-layer [CLS] representation, while post-submission experiments explored intermediate-layer representations and domain-category embeddings to improve generalization. For \textit{Task 2.1} and \textit{Task 2.2}, they adopted a separate detection-and-verification pipeline, using CamelBERT for hallucination detection and a QLoRA-tuned AceGPT-v2-8B-Chat to select the correct answer by directly comparing the logits of the six candidate options.

% \noindent
% \textbf{DzairVerse} fine-tuned a single Fanar-1-9B-Instruct model using QLoRA under a leak-free two-stage setup. The first stage is a hallucination detector that uses only the question and the generated answer, without exposing candidate options, while the second stage is answer selection, which subsequently uses the question, the generated answer, and six candidates to predict the correct option. The same QLoRA-adapted model performs both stages through separate prompts, with completion-only supervision used during fine-tuning.

\noindent
\textbf{Hashtag AI} proposed a QLoRA-adapted Gemma 4 12B system for hallucination detection and answer selection. Their approach uses a knowledge-graph retrieval system as an offline teacher to identify difficult facts and construct contrastive training examples, which are then distilled into the model during fine-tuning. The submitted system uses neither retrieval nor external knowledge at inference time. The system first predicts the hallucination label and then uses a second prompt with the six candidate answers to select the correct option.

% \noindent
% \textbf{HelsinkiTeam} participated in \textit{Task 2.2} and proposed a domain-adapted ALLaM-7B-Instruct model fine-tuned using LoRA. They augmented the official training data with approximately 30,000 domain-specific examples, including controlled factual corruptions, Qur’anic and Hadith texts, Islamic inheritance and law examples, and broader synthetic knowledge examples. The resulting model jointly performs hallucination detection and correct-answer selection as an instruction-following generation task, producing structured JSON outputs through greedy decoding.

%\noindent
%\textbf{IslamicAI}

% \noindent
% \textbf{ShadowDz} participated in \textit{Task 2.1} and fine-tuned Fanar-1-9B-Instruct using LoRA. Their approach employs a two-step inference pipeline designed to avoid textual-similarity shortcuts. The model first predicts whether the generated answer is hallucinated using only the question and response, without seeing the candidate options, and then receives the predicted label and six options to select the correct answer. Both stages use independently designed prompts and greedy decoding, with the second stage conditioned on the hallucination decision produced by the first.

\noindent
\textbf{NAMAA} proposed a CAMeLBERT cross-encoder that jointly encodes the question and generated answer and uses the hallucinated-class softmax probability as the prediction score. The team also investigated reference-aware modeling, knowledge distillation, LoRA-InfoNCE contrastive learning, model ensembling, temperature scaling, and MC-Dropout uncertainty estimation to analyze cross-question and distributional generalization, calibration, and selective prediction.

% \noindent
% \textbf{RaghadAlrasheed} participated in \textit{Task 1.1} and \textit{Task 1.2} by fine-tuning Qwen3-8B model using QLoRA, after finding it more suitable for hallucination detection than Arabic-specialized models, such as ALLaM-7B, Yehia-7B, and Fanar-1-9B. Their approach jointly encodes the question, gold reference answer, and generated answer in a single prompt and performs binary hallucination detection using the next-token probability of the label token from a single forward pass rather than free-form generation. The system also incorporates implementation adjustments to handle Qwen3's reasoning-tag behavior and to provide the reference answer through the shared-task inference interface.

\noindent
\textbf{REGLAT} proposed an ensemble of three Arabic pretrained encoders: MARBERTv2, CAMeLBERT-Mix, and ARBERT. Each model was independently fine-tuned as a binary classifier using the question and generated answer as input, with Focal Loss to emphasize difficult examples. The final system combines the three models' hallucination probabilities through unweighted soft voting, with each encoder contributing equally to the final prediction.

\subsection{Results and Discussion} \label{sec:results}
In this section, we present and discuss results for all subtasks. Appendix~\ref {sec:baselines} provides the baseline results, while Appendix~\ref{sec:methodological_analysis} presents a detailed methodological analysis of the submitted systems.
\begin{table*}[!t]
\centering
\small
\begin{tabular}{clcc|cc}
\toprule
& & \multicolumn{2}{c|}{\textbf{Development}} 
& \multicolumn{2}{c}{\textbf{Test}} \\
\cmidrule(lr){3-4}
\cmidrule(lr){5-6}
\textbf{Rank} &
\textbf{Team} &
\textbf{AUC-ROC} &
\textbf{F1-Macro} &
\textbf{AUC-ROC} &
\textbf{F1-Macro} \\
\midrule

1 & REGLAT
& 94.13 & \textbf{87.04}
& \textbf{77.17} & \textbf{65.76} \\

\textcolor{blue}{2} &
\textcolor{blue}{Baseline\_ARBERT} &
\textcolor{blue}{92.05} &
\textcolor{blue}{85.07} &
\textcolor{blue}{76.39} &
\textcolor{blue}{62.54} \\

3 & AyahVerse
& 92.17 & 86.01
& 76.30 & 64.01 \\

4 & NAMAA
& 92.63 & 85.95
& 75.96 & 64.44 \\  % Abdessalam perhaps 61.44 
\textcolor{blue}{5} &
\textcolor{blue}{Baseline\_MARBERT} &
\textcolor{blue}{91.44} &  
\textcolor{blue}{83.75} &
\textcolor{blue}{74.66} &
\textcolor{blue}{62.48} \\

6 & Sense
& \textbf{94.16} & 86.95
& 74.61 & 62.43 \\

\textcolor{blue}{7} &
\textcolor{blue}{Baseline\_CAMeLBERT} &
\textcolor{blue}{92.20} &
\textcolor{blue}{85.67} &
\textcolor{blue}{74.59} &
\textcolor{blue}{65.70} \\

\textcolor{blue}{8} &
\textcolor{blue}{Baseline\_mBERT} &
\textcolor{blue}{91.14} &
\textcolor{blue}{83.57} &
\textcolor{blue}{70.14} &
\textcolor{blue}{60.84} \\

9 & Anhnamxtanh
& 79.42 & 73.32
& 66.01 & 59.41 \\

10 & Scalar\_NITK
& 81.48 & 78.35
& 64.92 & 62.98 \\

11 & RaghadAlrasheed
& 80.74 & 70.45
& 59.03 & 50.36 \\

\bottomrule
\end{tabular}
\caption{Results for subtask~1.1 (Generalize Across Questions) on the development and test sets. Baseline systems are shown in blue.}
\label{tab:task11_results}
\end{table*}

\begin{table*}[t]
\centering

\small
\begin{tabular}{clcc|cc}
\toprule
& & \multicolumn{2}{c|}{\textbf{Development}} 
& \multicolumn{2}{c}{\textbf{Test}} \\
\cmidrule(lr){3-4}
\cmidrule(lr){5-6}
\textbf{Rank} &
\textbf{Team} &
\textbf{AUC-ROC} &
\textbf{F1-Macro} &
\textbf{AUC-ROC} &
\textbf{F1-Macro} \\
\midrule

1 & NAMAA
& 75.80 & 66.11
& \textbf{76.70} & 66.34 \\

2 & REGLAT
& 79.21 & 69.34
& 74.50 & 65.07 \\

\textcolor{blue}{3} &
\textcolor{blue}{Baseline\_MARBERT} &
\textcolor{blue}{79.64} &
\textcolor{blue}{69.34} &
\textcolor{blue}{73.92} &
\textcolor{blue}{65.24} \\

4 & AyahVerse
& 74.92 & 63.24
& 72.99 & 61.35 \\

\textcolor{blue}{5} &
\textcolor{blue}{Baseline\_ARBERT} &
\textcolor{blue}{79.27} &
\textcolor{blue}{68.53} &
\textcolor{blue}{72.08} &
\textcolor{blue}{62.81} \\

6 & Sense
& \textbf{83.25} & \textbf{71.52}
& 71.96 & 48.34 \\

\textcolor{blue}{7} &
\textcolor{blue}{Baseline\_CAMeLBERT} &
\textcolor{blue}{76.09} &
\textcolor{blue}{66.77} &
\textcolor{blue}{70.08} &
\textcolor{blue}{62.00} \\

\textcolor{blue}{8} &
\textcolor{blue}{Baseline\_mBERT} &
\textcolor{blue}{73.65} &
\textcolor{blue}{62.38} &
\textcolor{blue}{68.70} &
\textcolor{blue}{59.26} \\

9 & Scalar\_NITK
& 60.86 & 64.38
& 67.97 & \textbf{67.21} \\

10 & Anhnamxtanh
& 64.64 & 59.48
& 62.29 & 58.49 \\

11 & RaghadAlrasheed
& 51.66 & 50.96
& 49.07 & 43.05 \\

\bottomrule
\end{tabular}
\caption{Results for subtask~1.2 (Generalize Across Models) on the development and test sets.}
\label{tab:task12_results}

\end{table*}

\begin{table}[t]
\centering

\small
\begin{tabular}{clcc}
\toprule
\textbf{Rank} &
\textbf{Team} &
\textbf{Assisted } &
\textbf{Blind } \\
\midrule

\multicolumn{4}{c}{\textbf{Task 2.1: Islamic Knowledge}} \\
\midrule
1 & Anhnamxtanh   & \textbf{88.2} & \textbf{85.9} \\
2 & Hashtag AI & 81.2 & 82.6 \\
3 & DzairVerse    & 78.8 & 75.8 \\
4 & IslamicAI   & 73.1 & 65.4 \\
5 & HelsinkiTeam  & 72.0 & 66.6 \\
6 & ShadowDz      & 71.5 & 62.2 \\
\textcolor{blue}{7} &
\textcolor{blue}{Baseline} &
\textcolor{blue}{66.2} &
\textcolor{blue}{62.8} \\
\midrule

\multicolumn{4}{c}{\textbf{Task 2.2: General Culture}} \\
\midrule
1 & Anhnamxtanh & \textbf{85.7} & \textbf{83.5} \\
2 & IslamicAI   & 79.6& 64.4 \\
3 & AyahVerse   & 77.6 & 63.3 \\
4 & Vescera     & 72.2 & 62.6 \\
\textcolor{blue}{5} &
\textcolor{blue}{Baseline} &
\textcolor{blue}{71.5} &
\textcolor{blue}{67.9} \\
\bottomrule
\end{tabular}
\caption{Official results for Task~2, including subtask~2.1 (Islamic Knowledge) and subtask~2.2 (General Culture). Scores correspond to the final Task~2 score ($\mathrm{Score}_{\mathrm{Task2}}$) and are reported as percentages.}
\label{tab:task2_results}

\end{table}

\subsubsection{Track 1 Results}
\noindent
\textbf{subtask 1.1}
Table~\ref{tab:task11_results} presents the results for subtask 1.1, where systems are evaluated on their ability to generalize to unseen questions. \textit{REGLAT} achieved the best overall performance according to the primary ranking metric, obtaining a test AUC-ROC of 77.17. It was followed closely by the \textit{ARBERT baseline} (76.39), \textit{AyahVerse} (76.30), and \textit{NAMAA} (75.96), indicating a relatively narrow performance gap among the highest-ranked systems. In contrast, the remaining submissions achieved test AUC-ROC scores ranging from 74.61 to 59.03.

A notable observation is the substantial discrepancy between development and test performance. For example, \textit{REGLAT} decreased from 94.13 AUC-ROC on the development set to 77.17 on the test set, while \textit{Sense} decreased from 94.16 to 74.61. A similar pattern is observed for the baselines, including \textit{CAMeLBERT}, which decreased from 92.20 to 74.59. Importantly, this decline should be interpreted in light of the different dataset compositions of the two splits. The development set combines instances from both \textit{HalluTruthQA} and \textit{HalluScore}, whereas the test set consists exclusively of \textit{HalluScore}. Unlike conventional QA collections, \textit{HalluScore} is specifically constructed around hallucination-prone questions designed to elicit unsupported or unreliable responses from LLMs. Consequently, the test set represents a more challenging distribution in which systems must distinguish hallucinations among questions intentionally designed to expose model weaknesses. The development--test gap therefore reflects not only generalization to unseen questions, but also robustness to a shift toward a more challenging hallucination-oriented question distribution. These findings suggest that models achieving high performance on the mixed development distribution may rely on patterns that transfer less effectively to systematically hallucination-prone questions.

\textbf{Subtask 1.2} evaluates a different form of robustness: whether a hallucination detector can generalize to answers generated by unseen LLMs. As shown in Table~\ref{tab:task12_results}, \textit{NAMAA} achieved the highest test AUC-ROC of 76.70, outperforming \textit{REGLAT} (74.50) and the strongest baseline, \textit{MARBERT} (73.92). \textit{NAMAA} therefore improved on the best baseline by 2.78 AUC-ROC points, representing a clearer improvement on the provided baselines than observed in subtask~1.1. \textit{AyahVerse} also remained competitive, achieving 72.99 AUC-ROC, while \textit{Sense} obtained 71.96 despite achieving the strongest development performance on AUC-ROC (83.25). The results of subtask 1.2 again reveal a pronounced development--test discrepancy. \textit{Sense} provides the clearest example: despite ranking first on both development metrics, its test AUC-ROC dropped by 11.29 points to 71.96. \textit{REGLAT} similarly decreased from 79.21 to 74.50, while the \textit{MARBERT} baseline decreased from 79.64 to 73.92. In contrast, \textit{NAMAA} showed an unusual degree of robustness, with its AUC-ROC increasing slightly from 75.80 on development to 76.70 on test. This suggests that \textit{NAMAA} learned representations or decision criteria that transferred more effectively across model-generated outputs. More broadly, the variation across systems indicates that hallucination detectors can be sensitive to the characteristics of the LLMs used to generate their training data, reinforcing the importance of explicitly evaluating cross-model transfer rather than assuming that a detector trained on one set of generators will generalize to unseen ones.

\subsubsection{Track 2 Results}

Table~\ref{tab:task2_results} reports the official results for Task~2.
As described in Appendix~\ref{sec:baselines}, the systems are evaluated
under two settings: with access to the candidate options (\textbf{assisted})
and without access to them (\textbf{blind}). Overall, most systems perform
better in the assisted setting. In this setting, the model receives not only
the question and generated answer, but also six candidate answers consisting
of one correct option and five closely related plausible distractors. These
candidates provide additional factual alternatives that may lead the model
to revise its hallucination judgment, particularly when the generated response
is plausible but contains a specific factual error. In contrast, the blind
setting requires hallucination classification based solely on the question
and the generated answer. Excluding the baseline, the mean assisted score is approximately $5.99\%$
higher than the mean blind score in Subtask~2.1 and $15.12\%$ higher in
Subtask~2.2.
\\
For \textbf{subtask~2.1} \textit{(Islamic Knowledge)}, \textit{Anhnamxtanh} achieved the best performance in both settings, scoring $0.8824$ under assisted evaluation and $0.8593$ under blind evaluation. The system uses a QLoRA-adapted \textit{ALLaM-7B} model with a cross-check verification strategy. The narrow gap of $0.0231$ indicates that its performance remains stable without candidate information during hallucination detection. \textit{Hashtag AI} reached $0.8124$ in the assisted setting and a slightly higher $0.8262$ in the blind setting, making it the only system for which candidate access did not improve the score. Its \textit{Gemma 4 12B} model was trained with contrastive examples derived from an offline knowledge-based teacher, which may contribute to its strong blind performance. \textit{DzairVerse} also showed relatively stable results, with scores of $0.7881$ and $0.7582$, respectively. Its \textit{Fanar-1-9B-Instruct} system explicitly separates hallucination detection from answer selection, which is consistent with the limited degradation observed in the blind setting.
\\
For \textbf{subtask 2.2} \textit{(General Culture)}, \textit{Anhnamxtanh} again ranked first, obtaining $0.8565$ in the assisted setting and $0.8348$ in the blind setting. The small gap of $0.0217$ follows the same pattern as in subtask~2.1 and indicates stable performance across both evaluation settings and domains. In contrast, the remaining systems show larger assisted--blind gaps. \textit{AyahVerse} obtained $0.7762$ and $0.6331$, respectively. The latter combines \textit{CamelBERT} for hallucination detection with \textit{AceGPT-v2-8B-Chat} for answer selection. \textit{Vescera}, based on \textit{ALLaM-7B} with privileged-information distillation, obtained $0.7224$ in the assisted setting and $0.6256$ in the blind setting. These larger gaps indicate that candidate information provides a stronger signal for these systems during factual verification.

Overall, the two settings capture complementary aspects of factual verification. Assisted evaluation measures the benefit of candidate information, while blind evaluation provides a stricter assessment of hallucination detection. The results show that candidate answers improve detection, though some systems remain stable without them, highlighting the challenge of jointly detecting hallucinations and recovering correct factual information.

% Overall, the two settings capture complementary aspects of factual verification. Assisted evaluation measures the ability to exploit candidate answers as additional factual context, whereas blind evaluation provides a stricter assessment of hallucination detection directly from the question--response pair. The consistently small gaps achieved by \textit{Anhnamxtanh} indicate greater stability without candidate information, while the larger gaps observed for several other systems show that candidate answers can materially affect hallucination classification. This distinction is important because Task~2 goes beyond standard multiple-choice QA: systems must both detect hallucinated responses and recover the correct factual information among closely related alternatives.

% \section{Overview of Submitted Systems}\label{subsec:papers-desc}
% In this section, we present an overview of the submitted systems for each subtask and summarize the methodological approaches adopted by the participating teams.

% \textcolor{red}{Aisha and Dr. Abdessalam's team}

\section{Conclusion}\label{sec:conclusion}
 
We presented \textit{HalluScoring 2026}, a shared task for evaluating hallucination detection and factual verification in Arabic QA across unseen questions, unseen LLMs, and different knowledge domains. The results demonstrate that generalization remains challenging, with several systems showing substantial performance drops under distribution shifts, while Task 2 further highlights the distinction between detecting hallucinations and identifying the correct factual information. Overall, the findings emphasize the need for Arabic hallucination detection systems that are robust across questions and model generators and that can move beyond detection toward reliable factual verification.
\section*{Limitations \& Ethical Considerations}
\label{sec:limit}

\textit{HalluScoring 2026} has several limitations. The shared task focuses on Arabic QA and therefore does not directly assess whether the observed findings generalize to other languages or generation tasks. Moreover, the datasets cover a finite set of domains, questions, and LLM-generated responses, and performance may vary for unseen domains, model families, or future LLMs. The shared-task restrictions excluded retrieval-augmented generation and external search during inference. In Task~2, systems nevertheless received candidate answers during answer selection in both settings and during hallucination detection in the assisted setting.

From an ethical perspective, hallucination detection systems should not be interpreted as guaranteeing the factual correctness of LLM-generated content. This is particularly important in sensitive domains, including the religious content represented in the Islamic knowledge subtask, where incorrect quotations, attributions, or interpretations may have significant consequences. We therefore encourage the use of such systems as supporting tools rather than replacements for appropriate human or domain-expert verification.

\section*{Acknowledgments}

Aisha Alansari and Hamzah Luqman gratefully acknowledge King Fahd University of Petroleum and Minerals (KFUPM) for its support of this work. Muhammad Abdul-Mageed acknowledges support from Canada Research Chairs (CRC), the Natural Sciences and Engineering Research Council of Canada (NSERC; RGPIN-2026-07098), the Social Sciences and Humanities Research Council of Canada (SSHRC; 895-2020-1004), and Canadian Foundation for Innovation (CFI; 37771).

\normalem
\bibliography{refereces}

\appendix
\newpage
\section*{Appendix}
\label{sec:appendix}

\section{Annotation Guidelines}
\noindent
\subsection{HalluScoring Annotation Guidelines}
\label{sec:HalluScoringAnnotation}
The LLM responses were manually annotated by two annotators following a predefined hallucination annotation protocol. The annotation distinguished hallucinations from general errors. For instance, an unsupported or fabricated real-world claim presented as factual was considered a hallucination, whereas an incorrect response accompanied by uncertainty, hesitation, or an explicit indication of insufficient knowledge (e.g., ``I don't know'') was treated as an error rather than a hallucination. The annotations were performed against the verified ground-truth answers and supporting evidence provided in \textit{HalluScore}. Inter-annotator agreement (IAA) was measured using Cohen's $\kappa$ separately for each evaluated LLM's responses. The agreement was consistently high, ranging from $\kappa=0.9400$ to $\kappa=0.9768$ across the 17 models, corresponding to almost perfect agreement and demonstrating the high reliability of the human annotations.

\subsection{HalluTruthQA Annotation Guidelines}
\label{sec:HalluTruthQAAnnotation}
The annotation process was conducted in two stages. First, four domain
experts, one for each domain, verified the questions and reference answers
and constructed and validated the candidate answers and corresponding answer
keys. In a second stage, two trained research assistants independently
reviewed the questions, reference answers, candidate options, and answer
keys. Any unclear or disputed instance was returned to the corresponding
domain expert for final adjudication, providing an additional quality-control
step before the final dataset release. Inter-annotator agreement was high
across all annotation dimensions, with Cohen's $\kappa$ values of 0.92 for
the binary hallucination label, 0.86 for macro-level hallucination types,
and 0.80 for micro-level hallucination types.
\\
A response was labeled as \textit{hallucinated} if it contained at least one
factually incorrect, fabricated, unsupported, misleading, or contextually
inconsistent claim. Annotators evaluated the complete generated response
rather than considering only its main conclusion. Therefore, an answer could
be labeled as hallucinated even when its main conclusion was correct, but its
supporting evidence, numerical value, date, quotation, citation, or source
attribution was incorrect. A response was considered
\textit{non-hallucinated} when all factual claims relevant to the question
were correct and adequately supported.
\\

% \section{Data Samples}

% \textcolor{red}{Ahmed}

\section{Data Samples}
\label{sec:data_samples}

To illustrate the data released to participants, Tables~\ref{tab:sample_track1}
and~\ref{tab:sample_track2} present representative annotated examples from
Track~1 and Track~2, respectively. Each Arabic field is shown together with an
English gloss. The Track~1 sample follows the question--answer format used in
subtasks~1.1 and~1.2, where systems predict a binary hallucination label for a
generated answer. The Track~2 sample follows the format used in subtasks~2.1
and~2.2, which additionally provides a set of candidate answers with an answer
key and, for hallucinated responses, span-level annotations with explanations.

\begin{table}[t]
    \centering
    \includegraphics[width=\linewidth]{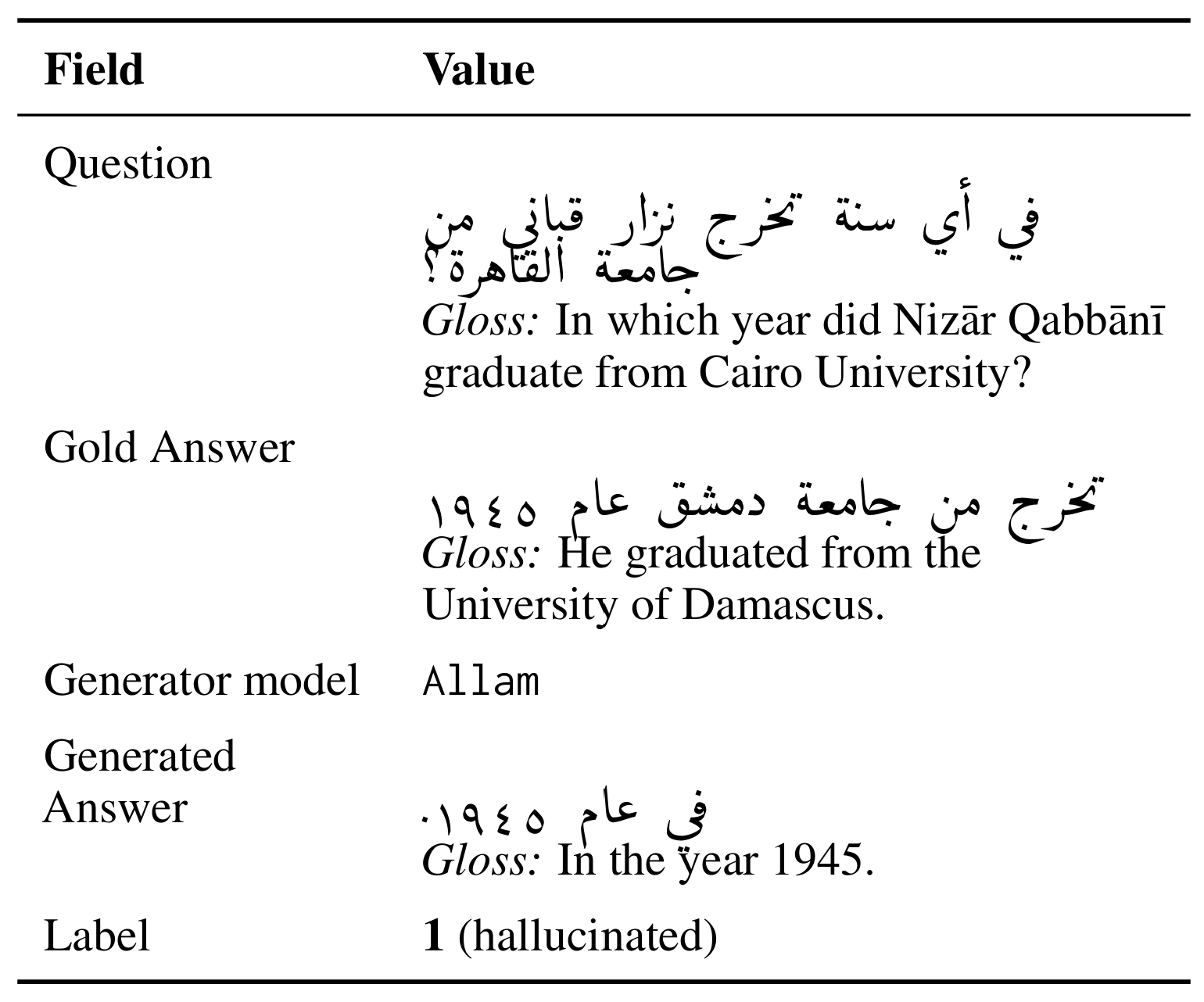}
    \caption{Representative annotated data sample provided to participants in
Track~1 (subtasks~1.1 and~1.2). Systems receive the question, the gold answer,
the generator model, and the generated answer, and must predict the binary
hallucination Label (1 = hallucinated, 0 = not hallucinated).}
    \label{tab:sample_track1}
\end{table}

\begin{table*}[t]
    \centering
    \includegraphics[width=\linewidth]{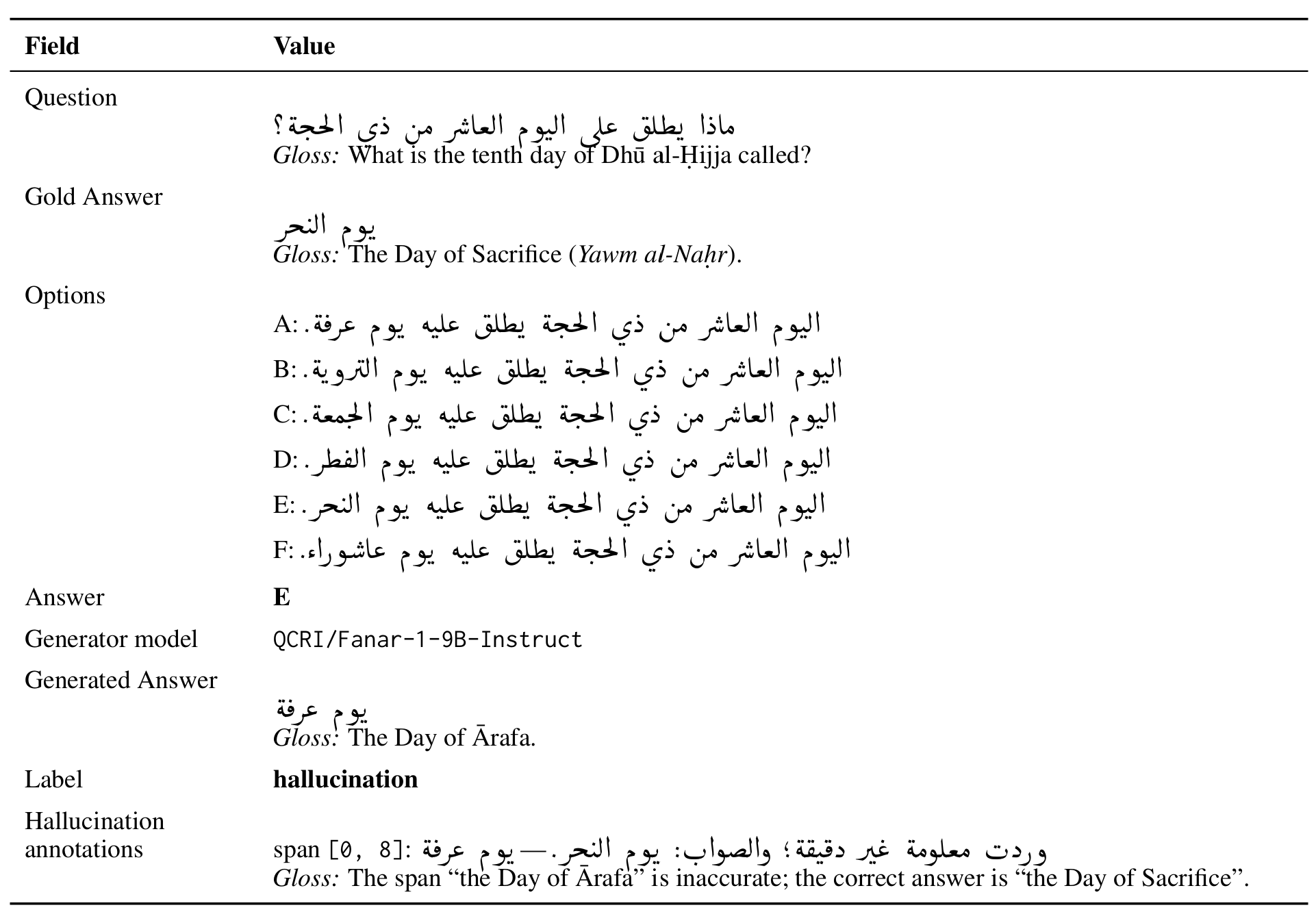}
    \caption{Representative annotated data sample provided to participants in
Track~2 (subtasks~2.1 and~2.2). Each instance
provides a set of candidate answers (Options) with an Answer
key, and, when the response is hallucinated, span-level Hallucination
annotations with explanations.}
    \label{tab:sample_track2}
\end{table*}

\section{Baseline Models}
\label{sec:baselines}
\noindent
\textbf{Task 1.} We provided the participants with baseline results established using four pretrained language models: CAMeLBERT, ARBERT, MARBERT, and multilingual BERT (mBERT).
Each model was fine-tuned on the training split and evaluated on the development set and the held-out test set. Moreover, we released the corresponding baseline code as a starter toolkit to facilitate system development and provide participants with a reproducible implementation for data loading, model fine-tuning, inference, and submission generation. 

For both subtasks~1.1 and~1.2, the question and generated answer were jointly provided to the classifier as a sentence pair, following the input format \texttt{[CLS] question [SEP] model\_answer [SEP]}. The input sequences were truncated to a maximum length of 512 tokens. All baseline models were fine-tuned using a batch size of 16, a learning rate of $2\times10^{-5}$, a linear warmup ratio of 0.1, and a weight decay of 0.01. Training was performed for up to 5 epochs, with early stopping if the development AUC-ROC did not improve for 3 consecutive epochs. For each model, the checkpoint achieving the highest development AUC-ROC was retained for evaluation on the test set. 

Among the four baseline systems, as shown in Table \ref{tab:task11_results}, ARBERT achieved the highest test AUC-ROC of 76.39, followed by MARBERT at 74.66, CAMeLBERT at 74.59, and mBERT at 70.14 in subtask 1.1. On the other hand, MARBERT served as the strongest baseline for Task~1.2 according to both evaluation metrics. The reduction in performance relative to the development results also highlights the difficulty of generalizing hallucination detectors to outputs produced by unseen language models.

\noindent
\textbf{Task 2.} We use the same model and generation configuration as those
used to produce the \texttt{generated\_answer} field
(see Section~\ref{HalluTruthQA_Dataset}), but with a different prompt and
a reduced maximum generation length from 1,024 to 256 new tokens.

Baseline prediction follows a two-stage process. In the first stage, the
model predicts whether the generated answer is hallucinated or not. In the
\textit{blind setting}, the model receives only the question and the generated
answer. In the \textit{assisted setting}, the six candidate answers are also
provided during hallucination detection.
\\
This baseline evaluates Fanar's ability to detect errors in its own generated
answers and the benefit of providing candidate answers for self-verification.

\section{Methodological Analysis}
\label{sec:methodological_analysis}
 
\begin{table*}[t!]
\centering

\small
\setlength{\tabcolsep}{4.5pt}
\renewcommand{\arraystretch}{1.15}

\begin{tabular}{lcc|cc|ccccc}
\toprule
& \multicolumn{2}{c|}{\textbf{Subtask}} 
& \multicolumn{2}{c|}{\textbf{Model Family}}
& \multicolumn{5}{c}{\textbf{Approach}} \\
\cmidrule(lr){2-3}
\cmidrule(lr){4-5}
\cmidrule(lr){6-10}

\textbf{Team} &
\textbf{1.1} &
\textbf{1.2} &
\textbf{Encoder} &
\textbf{Decoder} &
\textbf{Ensemble} &
\textbf{Reference} &
\textbf{Distill.} &
\textbf{Ext. Feat.} &
\textbf{PEFT} \\
\midrule

AyahVerse
& \checkmark & \checkmark
& \checkmark & 
& 
& 
& 
& 
& \\

NAMAA
& \checkmark & \checkmark
& \checkmark &
&
&
&
&
&
\\

RaghadAlrasheed
& \checkmark & \checkmark
& & \checkmark
&
& \checkmark
&
&
& \checkmark \\

REGLAT
& \checkmark & \checkmark
& \checkmark &
& \checkmark
&
&
&
& \\

Scalar\_NITK
& \checkmark & \checkmark
& \checkmark & \checkmark
& \checkmark
&
& \checkmark
& \checkmark
& \checkmark \\

Sense
& \checkmark & \checkmark
& \checkmark &
& \checkmark
&
&
&
& \\

\bottomrule
\end{tabular}
\caption{Overview of the approaches adopted by participating teams in 
Tasks~1.1 and~1.2. A checkmark indicates that the corresponding component 
was used in the submitted system. Distill.: knowledge distillation; 
Ext. Feat.: external features; PEFT: parameter-efficient fine-tuning.}
\label{tab:task1_system_overview}

\end{table*}

\begin{table*}[t!]
\centering

\small
\setlength{\tabcolsep}{2.5pt}
\renewcommand{\arraystretch}{1.15}

\begin{tabular}{lcc|cc|ccccccc}
\toprule
& \multicolumn{2}{c|}{\textbf{Subtask}}
& \multicolumn{2}{c|}{\textbf{Model Family}}
& \multicolumn{7}{c}{\textbf{Approach}} \\
\cmidrule(lr){2-3}
\cmidrule(lr){4-5}
\cmidrule(lr){6-12}

\textbf{Team} &
\textbf{2.1} &
\textbf{2.2} &
\textbf{Encoder} &
\textbf{Decoder} &
\textbf{PEFT} &
\textbf{Two-Stage} &
\textbf{Cross-Check} &
\textbf{Distill.} &
\textbf{Data Aug.} &
\textbf{Retrieval} &
\textbf{Priv. Info.} \\
\midrule

Anhnamxtanh
& \checkmark &
& & \checkmark
& \checkmark
&
& \checkmark
&
& \checkmark
&
\\

AyahVerse
& \checkmark & \checkmark
& \checkmark & \checkmark
& \checkmark
& \checkmark
&
&
&
&
\\

DzairVerse& \checkmark &
& & \checkmark
& \checkmark
& \checkmark
&
&
&
&
\\

Hashtag AI
& \checkmark & \checkmark
& & \checkmark
& \checkmark
& \checkmark
&
& \checkmark
& \checkmark
& \checkmark
& \\

HelsinkiTeam
& & \checkmark
& & \checkmark
& \checkmark
&
&
&
& \checkmark
&
\\

ShadowDz
& \checkmark &
& & \checkmark
& \checkmark
& \checkmark
&
&
&
&
\\

Vescera
& & \checkmark
& & \checkmark
& \checkmark
& \checkmark
&
& \checkmark
& \checkmark
&
& \checkmark \\

\bottomrule
\end{tabular}
\caption{Overview of the approaches adopted by participating teams in Tasks~2.1 and~2.2.
A checkmark indicates that the corresponding component was used in the submitted system.
PEFT: parameter-efficient fine-tuning; Distill.: knowledge distillation; 
Data Aug.: data augmentation; Priv. Info.: privileged information.}
\label{tab:task2_system_overview}

\end{table*}

\subsection{Task 1 Systems}
Table~\ref{tab:task1_system_overview} summarizes the main methodological choices adopted by participating teams in subtasks~1.1 and~1.2. A clear trend is the dominance of \textbf{encoder-based architectures}, which are provided in the starter toolkit. Five of the six teams employed encoder models, while decoder-based models were used only by \textit{RaghadAlrasheed} and \textit{Scalar\_NITK}. This preference is consistent with the discriminative nature of Task~1, where systems classify whether a given question--answer pair contains hallucinated information rather than generate new text. Moreover, all teams participated in both subtasks, allowing the same general modeling strategies to be evaluated under both question-level and cross-model generalization settings.

Another recurring strategy was \textbf{model ensembling}. \textit{REGLAT}, \textit{Sense}, and \textit{Scalar\_NITK} combined predictions from multiple models, suggesting that participants explored model diversity as a means of improving robustness. Notably, \textit{REGLAT}, which achieved the highest test AUC-ROC in subtask~1.1, used an encoder-based ensemble, while \textit{Sense} also adopted an encoder ensemble and obtained the strongest development AUC-ROC in both subtasks~1.1 and~1.2. Nevertheless, ensembling did not consistently guarantee stronger test generalization: \textit{NAMAA}, which achieved the highest test AUC-ROC in subtask~1.2, relied on an encoder-based approach without an ensemble. This suggests that effective cross-model generalization may depend more strongly on the learned representations and training strategy than on combining multiple classifiers alone.

More specialized techniques were less commonly explored. \textit{RaghadAlrasheed} employed a decoder-based architecture together with knowledge distillation and parameter-efficient fine-tuning (PEFT), whereas \textit{Scalar\_NITK} adopted the most heterogeneous methodology, combining encoder- and decoder-based models with ensembling, distillation, external features, and PEFT. Despite their greater methodological complexity, these systems did not outperform the simpler encoder-based approaches on the primary test metric. Overall, the submitted systems indicate that \textbf{pretrained encoder representations, optionally combined through ensembling, remain a strong paradigm for Arabic hallucination detection}. At the same time, the limited use of reference-based information, contrastive learning, and other explicit factuality-oriented mechanisms reveals opportunities for future work, particularly for improving robustness to the distribution shifts and unseen model outputs considered in subtasks~1.1 and~1.2.

\subsection{Task 2 Systems}

Table~\ref{tab:task2_system_overview} summarizes the main methodological choices adopted by participating teams in subtasks~2.1 and~2.2. In contrast to Task~1, \textbf{decoder-based architectures dominated Task~2}, reflecting the more complex nature of the task, which requires systems not only to detect hallucinated responses but also to identify the correct factual answer. All participating teams employed decoder-based models, with \textit{AyahVerse} additionally incorporating an encoder-based component. Parameter-efficient fine-tuning (PEFT) was also widely adopted across the submitted systems, indicating a preference for adapting pretrained language models to the task without requiring full-model fine-tuning.

Several teams employed \textbf{multi-stage or verification-oriented strategies}. 
\textit{AyahVerse}, \textit{DzairVerse}, \textit{Hashtag AI}, \textit{ShadowDz}, 
and \textit{Vescera} used two-stage approaches, separating different components 
of the prediction process rather than treating the task as a single classification 
decision. Cross-checking mechanisms were used by \textit{Anhnamxtanh}, 
\textit{DzairVerse}, \textit{ShadowDz}, and \textit{Vescera}, reflecting an effort 
to explicitly verify intermediate or final predictions. Other strategies were less 
common. \textit{Hashtag AI} used knowledge distillation, while 
\textit{Anhnamxtanh} employed counterfactual supervision and cross-role agreement. 
Moreover, \textit{Hashtag AI} and \textit{Vescera} adopted data augmentation. 
Retrieval was used by \textit{Hashtag AI} and \textit{HelsinkiTeam}, providing 
additional information that could support factual verification and correct-answer 
identification. \textit{Vescera} additionally incorporated privileged information, 
making it the only submitted system to explore this strategy.

The methodological diversity observed in Task~2 highlights the additional challenges introduced by factual verification compared with binary hallucination detection. The submitted systems demonstrate that no single methodological component characterizes all high-performing approaches. Overall, Task~2 encouraged greater use of \textbf{generative models, staged reasoning, and explicit factual verification mechanisms} than Task~1, reflecting the need to move beyond identifying whether a response is hallucinated toward determining what the correct factual information should be.

\end{document}